\documentclass[runningheads]{llncs}
\usepackage{graphicx}
\usepackage{array}
\usepackage{amsmath}
\usepackage{algorithm}
\usepackage{algorithmicx}
\usepackage{algpseudocode}
\usepackage{graphicx}
\usepackage{multirow}
\usepackage{makecell}
\usepackage{color}
\usepackage{booktabs}
\usepackage{cite}
\usepackage{subcaption}
\usepackage{marvosym}

\begin{document}
\title{Semantic-based Data Augmentation for Math Word Problems}
%
%\titlerunning{Abbreviated paper title}
% If the paper title is too long for the running head, you can set
% an abbreviated paper title here
% %
% \author{Ailisi Li \and Jiaqin Liang \and Yunwen Chen \and Yanghua Xiao}
% \authorrunning{A. Li et al.}
% % \authornote{Both authors contributed equally to this research.}
% \institute{Fudan University \\
% \email{\{alsli19,shawyh\}@fudan.edu.cn} \\ 
% \email{l.j.q.light@gmail.com}}
% % \affiliation{
% %   \institution{Fudan University}
% % }

% \renewcommand{\thefootnote}{\fnsymbol{footnote}}

\author{Ailisi Li \inst{1}  \and
Yanghua Xiao\inst{1 \and 2}\thanks{Corresponding author.} 
\and Jiaqing Liang \inst{1}
\and Yunwen Chen\inst{3}}
\authorrunning{A. Li et al.}

\institute{Shanghai Key Laboratory of Data Science, School of Computer Science, Fudan University, Shanghai, China\\
\and
Fudan-Aishu Cognitive Intelligence Joint Research Center, Shanghai, China
\and
DataGrand Inc., Shanghai, China\\
\email{\{alsli19,shawyh\}@fudan.edu.cn} \\ \email{l.j.q.light@gmail.com}, 
\email{chenyunwen@datagrand.com}}

% \footnotetext[1]{Corresponding author.}

\maketitle              % typeset the header of the contribution
\begin{abstract}
	It's hard for neural MWP solvers to deal with tiny local variances. 
	In MWP task, some local changes conserve the original semantic while the others may totally change the underlying logic. 
	Currently, existing datasets for MWP task contain limited samples which are key for neural models to learn to disambiguate different kinds of local variances in questions and solve the questions correctly.
	In this paper, we propose a set of novel data augmentation approaches to supplement existing datasets with such data that are augmented with different kinds of local variances, and help to improve the generalization ability of current neural models.
	New samples are generated by knowledge guided entity replacement, and logic guided problem reorganization. The augmentation approaches are ensured to keep the consistency between the new data and their labels. 
	Experimental results have shown the necessity and the effectiveness of our methods. 

\keywords{Math word problem  \and Data augmentation}
\end{abstract}

\section{Introduction}\label{section:intro}
% def and move1a
Automatically solving Math Word Problem (MWP) has attracted more and more research attention in recent years. The MWP solvers are fed in with a natural language description of a mathematical question, and output a solution equation as the answer. In most cases, these questions are short narratives comprised of several known quantities and a query about an unknown quantity, whose value is the answer we desire. Table \ref{exm} shows a typical example of MWP, where $x$ in the equation refers to the unknown quantity, and is calculated from the known quantities and specific constants such as $\pi$, 1, 2.

% The question contains two clue sentences each with a known quantity (\textit{390} and \textit{40\%} respectively), and a query with an unknown variable (\textit{kilograms of apples}). 

  %One of representative MWP is single equation problem, which can be solved by one equation with only basic arithmetic operators. A single equation problem is expressed by a natural language question.

% overview of previous work and their limitations
Numerous efforts have been devoted to solving this challenging task. 
Early studies relying on hand-crafted features \cite{roy2016solving, liang2016tag, roy2016unit} and predefined patterns \cite{shi2015automatically} have limitations in generalization. %, consequently performing poorly on larger datasets.
Deep learning methods have become popular to solve the MWP task in recent years \cite{wang2017deep, wang2019template,xie2019goal, zhang2020graph} due to their better capability of generalization. %they are good at generalization.
\cite{wang2017deep} first modeled the MWP task as an equation generation task, and various works have followed this framework since then.
Recent works in MWP mostly focus on designing complex generation models to capture more features from limited data. 
For example, \cite{xie2019goal} proposed a tree-structured decoder to imitate human behaviour, \cite{zhang2020graph} utilized GCN, and \cite{li2019modeling} adopted the attention mechanism to better capture the relationship between quantities.
%However, the performances of these deep methods highly rely on the quality and the scope of labeled training data. 

\begin{table}[t]
	\normalsize
	\centering
	\small
	\caption{An example of math word problem and new samples generated by semantic-based data augmentation approaches.}
		\begin{tabular}{|m{8.5cm}|}
			\hline
			\textbf{Original} \\
			\hline
			% \hline
			\textbf{Question 1:} There are \textit{390} kilograms of \textcolor{red}{pears} in the \textcolor{red}{store}, which is \textcolor{red}{\textit{40\%}} less than the weight of \textcolor{red}{apples}. \textcolor{red}{$x$} kilograms of \textcolor{red}{apples} are there in the \textcolor{red}{store}. \\
			% \hline
			\textbf{Equation 1:} $x=390\div(1-40\%)$ \\
			% \hline
			\textbf{Answer 1:} 650 \\
			\hline
			\textbf{Knowledge guided entity replacement} \\
			\hline
			% \hline
			\textbf{Question 2:} There are \textit{390} kilograms of \textcolor{red}{\textit{bananas}} in the \textcolor{red}{\textit{kitchen}}, which is \textit{40\%} less than the weight of \textcolor{red}{\textit{watermelon}}. $x$ kilograms of \textcolor{red}{\textit{watermelon}} are there in the \textcolor{red}{\textit{kitchen}}.\\
			% \hline
			\textbf{Equation 2:} $x=390\div(1-40\%)$ \\
			% \hline
			\textbf{Answer 2:} 650\\
			\hline
			\textbf{Logic guided problem reorganization} \\
			\hline
			% \hline
			\textbf{Question 3:} There are \textit{390} kilograms of pears in the store, which is \textcolor{red}{$x$} less than the weight of apples. \textcolor{red}{\textit{650}} kilograms of apples are there in the store.\\
			% \hline
			\textbf{Equation 3:} $x=1-390\div650$ \\
			% \hline
			\textbf{Answer 3:} 0.4\\
			\hline
		\end{tabular}
		\label{exm}
\end{table}

However, current MWP solvers still have weaknesses in terms of robustness and generalization. A superior MWP solver should understand a problem precisely in two ways. First, it is able to generate the same equation for a transformed question with only uninfluential entity replacement. For example, the equation generated for $Q1$ in Table \ref{exm} should not be changed when the \textit{pears} in the question is replaced with \textit{bananas}.
Second, excellent models should be capable of generating a different equation when the logic of the question changes even if \textbf{the text of the question only changes slightly}. 
For example, in Table \ref{exm}, the only difference between $Q1$ and $Q3$ is the position of token $x$. But their underlying logic is totally different and thus the corresponding equations are different. 
% For example, as shown in Table \ref{exm}, the question in the first box and the question in the last one are different only in the position of token $x$, but the underlying logic is totally different and thus the equations generated for them should be different.
These two kinds of tiny local variances in questions lead to totally different results, one of which conserves the underlying logic, while the other one changes it completely. %and the equation should be converted respectively.
Humans are able to disambiguate these local variances easily while it's hard for most neural models to deal with discrete local variances. Previous works on MWP hardly consider this challenge. 
% It is still challenging for current neural MWP solvers to disambiguate such local variances and make correct inference. %especially without massive training data.

The limitation of existing datasets for MWP is a main reason for the above-mentioned weaknesses.
% Existing datasets are either too small or less challenging. %contain few challenging samples 
Since labeling MWP data is time-consuming, existing MWP datasets are all too small compared to datasets for other natural language processing tasks. 
Besides, few challenging samples with similar questions but different equations can be found in MWP datasets, which makes it hard for neural models to learn to deal with tiny local variances.
% A challenging dataset should contain a significant ratio of samples with questions fairly similar with each other but different solution equations. %\textcolor{red}{same solution equation?}
% (challenging samples refer to the question pairs that are fairly similar in text, and the equations of them are either the same or not depending on the type of the local variances as described above).
% Datasets such as AI2 \cite{hosseini2014learning}, SingleEQ \cite{koncel2015parsing}, and AllArith \cite{roy2016unit} only have hundreds of samples. 
The most popular and largest single-equation MWP dataset Math23k contains only 23,161 problems, which is rather limited compared to datasets in other field like sQuAD \cite{rajpurkar2016squad} with 150,000 questions. 
% Besides, the amount and the quality of the challenging samples it contains are also not satisfactory as analyzed in \ref{section:training_set_analysis} and \ref{section:quality_of_challenging samples}.
% While relatively large-scale MWP datasets such as MAWPS \cite{koncel2016mawps} is constructed with low lexical and template overlap, which suggests limited coverage of challenging samples. %amount of challenging samples. 
% Math23k \cite{wang2017deep}, another popular large-scale MWP dataset, also contains a very limited amount of challenging samples and the quality of these samples are not quite satisfactory according to our experimental results (refer to Sec. \ref{section:training_set_analysis} and \ref{section:quality_of_challenging samples} for more details).
% Besides, even the largest single-equation MWP dataset Math23k contains only 23,161 problems, which is rather limited compared to other datasets like sQuAD with 150,000 questions \cite{rajpurkar2016squad}. 
The weaknesses of existing datasets, especially the limited coverage of challenging samples, motivated us to augment the dataset with questions of minor variances leading to heterogeneous equations.

In this paper, we propose a set of semantic-based data augmentation approaches suitable for MWP task, namely knowledge guided entity replacement and logic guided problem reorganization. And two kinds of local variances are provided accordingly. %to supplement existing datasets. 
Neural MWP solvers can benefit from our augmentation strategies in terms of generalization and the ability of dealing with tiny local variances.  
Unlike other popular augmentation approaches \cite{wei2019eda, xie2019unsupervised, yu2018qanet}, which may cause inconsistency of the questions and equations in MWP task, our augmentation methods are carefully designed for MWP task to ensure consistency.
% To the best of our knowledge, we are the first to employ data augmentation in MWP task. 
% Both augmentation methods are carefully designed to ensure the consistency of the questions and the equations while diversifying the training data in the meanwhile. 

\textbf{Knowledge augmentation (knowledge guided entity replacement).} 
As shown in Table \ref{exm}, \textit{pears} in the original question is replaced with another fruit \textit{bananas}. 
And it is obvious that the replacement does not change the original logic, thus the new question conserves the original label. 
Our knowledge guided entity replacement method randomly replaces several entities in questions with other entities that belong to the same concept as the original ones. 
And we guide the replacement with knowledge base that contains much taxonomy knowledge.
% And knowledge base containing much taxonomy knowledge is used to guide the replacement. 

Different from synonym replacement \cite{wei2019eda, fadaee2017data} which replaces random words with their synonyms, replacing entities could largely avoid semantic shifting since most entities in MWP are not crucial for the logic inference. 

\textbf{Logical augmentation (logic guided problem reorganization).} 
% Another way to realize the influential local disturbance is logic guided problem reorganization. 
% The other augmentation strategy we proposed is logic guided problem reorganization.
As the example shown in Table \ref{exm}, quantities \textit{390} and \textit{40\%} are known while quantity \textit{650} is unknown in $Q1$. 
In the generated $Q3$, we let \textit{40\%} be the unknown one given \textit{390} and \textit{650}. 
The equation is changed accordingly to keep consistency. $x$ in Eq 3 is substituted for \textit{650} and \textit{40\%} is replaced with $x$.
Afterwards, we transform the new equation to its equivalent equation of the form $x=1-390\div 650$. 
To enrich the problem types of the training data, our logical augmentation iteratively set the known quantities in the original question and equation to the unknown. %to keep the consistency. %And the equation is changed accordingly to keep the consistency. 
And the new equation is further transformed to its equivalent equation with mathematical properties for normalization. 

It is worth mentioning that previous works only learn the equation of the unknown, while our augmentation method helps the neural models to make full use of the limited data by learning all the possible equations of the quantities in the question.

Our contributions are summarized as follows:
\begin{itemize}
	\item To the best of our knowledge, this is the first systematical study of data augmentation for MWP task. And is easy to be applied to any neural models and extend to other math-related tasks.
	% \item We propose a set of novel semantic-based data augmentation methods which could be applied to any neural models for MWP to diversify the training data. And our methods are able to ensure the consistency of the data and the labels. %Our strategies can be applied to any neural models for MWP.
	\item Our methods can generate coherent questions with consistent labels, which largely diversify both textual descriptions and equation templates. It also brings in massive challenging samples that existing datasets lack.
% 	\item Our augmentation approaches largely diversify both textual descriptions and equation templates of the training data, and supplement existing datasets with massive challenging samples that they don't include currently. 
	% \item We propose a set of novel semantic-based data augmentation methods to diversify the training data. Our strategies are able to ensure the consistency of the data and the labels and can be applied to any neural models for MWP.
	% \item We propose a set of novel semantic-based data augmentation methods to diversify the training data that are able to ensure the consistency of the data and the labels. Our strategies can be applied to any neural models for MWP.
	\item Experimental results show the necessity and effectiveness of our methods. The performances of the additional evaluation also indicate that our methods largely enhances the generalization ability of neural models.
% 	that semantic-based augmentation largely enhances the generalization ability of neural models and remarkably boosts the performances of neural models.
	% neural models trained with data augmented by our methods are remarkably better than the baseline models. Even the simplest vanilla seq2seq model beats many complex models with our augmentation approaches. 
\end{itemize}
% \begin{itemize}
% \item Previous work mainly focuses on models while our work makes the first attempt to augment data in MWP. To the best of our knowledge, this is the first systematical study of data augmentation for MWP task.
% \item We propose a set of novel semantic-based data augmentation methods which are able to enhance the performances of different kinds of neural models for MWP task. 
% \item Experimental results show that neural models trained with data augmented by our methods are remarkably better than the baseline models. Even the simplest vanilla seq2seq model beats many complex models with our augmentation approaches. %In general, the proposed data augmentation strategy helps us achieve the state-of-the-art performance on Math23k \cite{wang2017deep}.
% \end{itemize}

% \input{intro.tex}

\section{Related Work}
% In this section, we'll introduce related work of MWP and data augmentation approaches applied in NLP tasks.
\subsection{Math Word Problem}
Early works mostly utilized statistical methods or predefined rules. 
\cite{roy2016solving} utilized hand-crafted features to predict the lowest common ancestor operator for each quantity pair. %pair of quantities mentioned in the text. %A classifier named significant number identification was used to reduce the tree enumeration space.
\cite{roy2016unit} proposed a unit dependency graph based on \cite{roy2016solving}.
\cite{liang2016tag} predefined a group of logic forms and converted the math question into them. %The solution was inferred from the logic forms. %Afterwards, the logic inference was performed to obtain the solution.

\cite{wang2017deep} first proposed to utilize a seq2seq model with recurrent neural network to generate equation template sequence.
% Deep neural solver proposed by \cite{wang2017deep} was the first deep learning based method for addressing MWP which utilized a seq2seq model with recurrent neural network to generate equation template sequence. 
\cite{wang2018translating} proposed equation normalization to unify duplicated representations of equivalent expressions. 
\cite{chiang2018semantically} utilized a seq2seq model with the help of stack to align the semantic with the operator.
\cite{wang2019template} proposed a two-stage algorithm to predict a template tree. 
\cite{li2019modeling} applied group attention mechanism to extract more features.

\cite{liu2019tree, xie2019goal, zhang2020graph} replaced the recurrent neural network based decoder with a tree-structured decoder and achieved satisfactory results. 
% \citet{xie2019goal} simulated human behaviour to decompose the goal into sub-goals in a top-down manner.
% \citet{zhang2020graph} introduced graphs to extract more features from original text and quantities.
\cite{zhangteacher} leveraged the framework of knowledge distillation.

% However, none of the works have tried to enhance the generalization ability of the neural models by data augmentation strategies. 
Above-mentioned works all focused on the model architectures while the weakness of lacking in labeled data has been ignored.
\cite{roda} is another augmentation work for MWP and is very similar to our logic guided augmentation method. However, since \cite{roda} re-order the sub-sentences of the original problem, it is likely to generate incoherent questions while our method avoid this weakness.
% \footnote{Our work was first submitted in Sep, 2020 which is }

\subsection{Data Augmentation for Natural Language Processing (NLP)}
We categorize data augmentation approaches for NLP into two types. The first type changes only the text while the second one changes both the text and labels.
% We categorize data augmentation approaches for NLP into three types. The first one is to add random noise to input data \cite{xie2017data} or hidden states \cite{le2015simple}.

Some research adds random noise to input data \cite{xie2017data} or hidden states \cite{le2015simple} making the models less sensitive to small perturbations.
\cite{wei2019eda} systematically examined some basic augmentation methods including random synonyms replacement, word insertion, etc.
\cite{xie2019unsupervised} utilized tf-idf to help determine which words to replace.
% only replaced unimportant words and keep keywords, determined by tf-idf, unchanged. 
% The second type of approaches performs local transformation on text, which solely changes a small part of text each time. Typical operations include word insertion, word deletion, synonyms replacement and exchange of word order. \cite{wei2019eda} systematically examine the benefit of these simple methods.
% Among them, synonyms replacement is the most widely-used local transformation approach. \cite{wei2019eda} just randomly selected words to be replaced with a random synonym of it, while \cite{xie2019unsupervised} only replaced unimportant words and keep keywords, determined by tf-idf, unchanged. 
\cite{wang2015s} adopted k-nearest neighbors to find synonyms in word embedding space. 
% The above augmentation methods change only the text. 
The Noise brought in by these methods could be tolerated in some tasks but not in MWP task due to the strict requirement of preciseness. 

\cite{kobayashi2018contextual, wu2019conditional} fine tuned a pre-trained language model with text and label to generate new sentences given specific labels.
% , and use words predicted by the language model given a masked sentence and a predefined label, to replace the original words.
\cite{fadaee2017data} replaced words in the source and target sentences with rare words.
% aimed at generating data for neural machine translation task in the low-resource situation, so that the new data was generated by 
Other approaches mainly applied generative models like VAE \cite{hu2017toward}, seq2seq model \cite{hou2018sequence}, GPT \cite{anaby2020not}, etc., to generate new text given a specific label.
All of the methods above generating new text given a specific label are inappropriate to MWP task, because the equation is a sequence and is not enumerable.
% The last one is global transformation which generates a new text with diverse textual expressions. The most popular approach must be back translation proposed in \cite{yu2018qanet}, whose idea was to utilize two machine translation models to translate original sentence from language A to language B, and then from B back to A.

\section{Methodology}
\subsection{Problem Statement} 
The MWP dataset contains training data $D_{train}$ and testing data $D_{test}$, both of which are comprised of numerous questions $Q$ and equations $Eq$. 
Neural models take $Q$ as the input and generate the $Eq$ sequence.
In previous works, the neural model $\mathcal{M}$ is trained with the training data $D_{train}$ and tested with the testing data $D_{test}$.
Due to the limitations of $D_{train}$ introduced in Sec. \ref{section:intro}, we generate new samples $D_{aug}$ from $D_{train}$ by our augmentation approaches. And the neural model $\mathcal{M}$ is trained with both original and augmented data $D_{train}\cup D_{aug}$.

The question $Q$ consists of a sequence of tokens $\mathcal{W}=\{w_i\}_{i=1}^{\left | \mathcal{W} \right |}$ including known quantities $\mathcal{N}=\{n_i\}_{i=1}^{\left | \mathcal{N}\right |}$ and entities $\mathcal{E}=\{e_i\}_{i=1}^{\left | \mathcal{E}\right |}$. %(we use entity to represent the mention in $Q$ for simplification). 
% Since quantities are fairly sparse and the neural models usually do not care about the exact values of them when solving a MWP, we replace quantities in $Q$ with a symbol $n_i$ ($i$ is determined by the sequence order) during preprocessing phase. Quantities in $Eq$ are replaced accordingly and we denote the answer of $Eq$ as $\hat{n}$
Since quantities are fairly sparse, we replace quantities in $Q$ with symbol $n_i$ according to the occurrence order of the quantities during preprocessing phase. 
Quantities in $Eq$ are replaced accordingly and we denote the answer of $Eq$ as $\hat{n}$

The target of MWP is to generate the $Eq$ sequence which is composed of $\mathcal{N} \cup \mathcal{O} \cup \mathcal{C}$. Among them, $\mathcal{O}$ is the set of the operators (such as $\times$) and $\mathcal{C}$ is the constant set.

\subsection{Knowledge Augmentation}\label{section:knowledge_aug}
The category information of entities is introduced to generate new questions. 
Given a sample ($Q$, $Eq$), we randomly choose $\theta$ entities mentioned in $Q$ to be replaced with other entities belonging to the same concept as the original ones. 
If an entity $e_i$ belonging to concept $c$ is selected, the alternative entities are $e_j \in \mathcal{E}_{c} (i \neq j)$. 
Similar to \cite{wei2019eda}, we set $\theta=\max(1, \alpha l)$, among which $l$ refers to the length of $Q$ and $\alpha$ is a hyper-parameter used to manage the replacement ratio.
More entities are replaced for longer questions considering they tolerate noise better. 
We notice that an entity may appear more than once in a question, and all of them should be replaced if the entity is selected. 
For each question $Q$, 
% we generate $N_{K}$ new questions with this strategy. 
the new questions generated by this means are denoted as $Q_K$. 
As an example in Table \ref{tab:knowledge_exm}, entities \textit{Ming Zhang} and \textit{apples} are replaced with a random entity of \textit{person} and \textit{fruit} accordingly.
Since entities are less informative for MWP inference, the generated questions conserve the logic of the original question. 
% And the new equations $Eq_K$ thus matched with $Q_K$ are still $Eq$.
And the new equation matched with $Q_K$ is still $Eq$.
\begin{table}[t]
	\small
	\centering
	\caption{An example of the question generated with knowledge augmentation.}
	\begin{tabular}{|m{1cm}<{\centering}|m{7cm}|}
	% \begin{tabular}{|c|l|}
		\hline
		$Q$ & \textit{Ming Zhang ([PER])} bought $n_0$ \textit{apples ([FRU])} \\
		\hline
		$Q_K$ & \textit{Hong Li ([PER])} bought $n_0$ \textit{blueberries ([FRU])}.\\
		\hline
	\end{tabular}
	\label{tab:knowledge_exm}
\end{table}

\subsubsection{Recognize entities.} 
There are two kinds of entities in MWP, namely \textbf{real-world entities} and \textbf{named entities}, and we take different strategies to recognize them. 
For detecting named entities like fake person names \textit{Alice, Bob}, etc. which are common in MWP, 
% Person names like \textit{Alice, Bob}, etc. are common in MWP. %, however, they are not real persons but a representation of the concept \textit{person} in MWP. 
% Replacing them with \textit{Mary} or \textit{John} does not make any difference. 
there are numerous models and tools work very well on this task. And we follow the method of \cite{che2020n} to recognize named entities.
% We adopt Named entity recognition is used to detect fake entities in questions.
% Fake entities in MWP are usually the names of persons, organizations, and locations. Named entity recognition (NER) is the best choice to recognize this kind of entity. 
Another kind of entities is real-world entities like \textit{apple, car}, etc. These entities can be easily recognized by referring to KGs. 
In this paper, we use WordNet \cite{miller1995wordnet} to guide our knowledge augmentation. 
Entities are linked to the WordNet syn-sets, and the direct hypernyms are viewed as their concepts. 
%Due to the ambiguity of natural language, a single phrase could refer to numerous entities and thus be mapped to several different WordNet synsets. 
%Considering in MWP scenario, the occurring entities are mostly popular entities (e.g apple in MWP mostly refers to the most common entity \textit{the fruit of apple} rather than the entity \textit{the corporation of Apple}). 
%Thus we simply select the most popular entity to disambiguate the phrases.
%Besides, we keep only \textit{physical entities} while ignoring the entities that belong to an abstract concept such as \textit{action} etc. to avoid possible semantic shifting.
Besides, we restrict that only \textit{physical entities} in questions could be replaced to avoid possible semantic shifting. \textit{Abstract entities} like \textit{time} should not be replaced in our case.

\begin{table}[t]
	\small
	\centering
	\caption{An example of the question generated with logical augmentation.}
	\begin{tabular}{m{1cm}<{\centering}|m{11cm}}
	\hline
		$Q$ & There are \textbf{$n_1$} kilograms of pears in the store, which is \textbf{$n_2$} less than the weight of apples. How many kilograms of apples are there in the store? \\
		\hline
		$Q^{'}$ & There are \textbf{$n_1$} kilograms of pears in the store, which is \textbf{$n_2$} less than the weight of apples. $x$ kilograms of apples are there in the store. \\
		\hline
		$Q_L$ & There are \textbf{$n_1$} kilograms of pears in the store, which is $x$ less than the weight of apples. \textbf{$\hat{n}$} kilograms of apples are there in the store. \\
	\hline
	\end{tabular}
	\label{exm_logical}
\end{table}

\subsection{Logical Augmentation}\label{section:logical_augmentation}
Given a sample ($Q$, $Eq$), several ($Q_L$, $Eq_L$) pairs are generated by setting a known quantity in the original question to the unknown one in the new question. 
And the new equation is normalized with mathematical properties. 
Since the textual description of $Q_L$ is similar to that of $Q$, neural models tend to generate similar equation sequences for them even though their ground truth solutions are totally different. 
With logical augmentation, neural models are forced to learn the different equations from the slight variations of input text, which to some extent enhances the inference ability of neural models.
\subsubsection{Question generation}
In MWP, the letter $x$ is usually assigned to represent the unknown. The value of it is the answer to the question $Q$ (denoted as $\hat{n}$). 
The unknown quantities in questions are usually indicated by question words such as \textit{how many}. 
Considering both question words and letter $x$ refer to the same unknown, we replace the question words in $Q$ with the letter $x$. 
And then the original question could be viewed as an assertive sentence about the logical relationship between the known and unknown quantities $\mathcal{N} \cup \hat{n}$.
Previous training without augmentation only learns how to get $\hat{n}$ given $\mathcal{N}$, for $Eq$ is in $x=f(\mathcal{N} \cup \mathcal{C})$ form, in which $x$ refers to $\hat{n}$ as $\hat{n}$ is the unknown of $Q$. %the unknown symbol $x$ refers to $n_{\left |\mathcal{N}\right |+1}$. 
However, the logical relationship between other quantity pairs remains unlearned. 
To make full use of each data sample, we make the neural models additionally learn how to get $n_i \in \mathcal{N}$ given the other quantities $n_j \in \mathcal{N} \cup \hat{n} (i \neq j)$.
As the letter $x$ indicates the quantity we desire, quantities in $Q$ are iteratively set to $x$. The generated questions are denoted as $Q_L$. 

Table \ref{exm_logical} shows an example of questions generated with logical augmentation. Notably, when training with augmented data, original questions $Q$ are replaced with $Q^{'}$ to ensure that all questions are expressed in an uniform form.
% As shown in Table \ref{exm_logical}, question word \textit{how many} in $Q$ is replaced with $x$ to get $Q^{'}$. 
% Afterwards, the originally known quantity $n_2$ in $Q^{'}$ is replaced with $x$ in the new question $Q_L$. As a result the unknown turns from $\hat{n}$ to $n_2$. 
% And the neural models will learn how to get $n_2$ given $\{n_1,\hat{n}\}$ with $Q_L$. 
% replace the given quantity $n_i$ with an unknown symbol $x$ and replace the original $x$ with the original answer value $n_{\mathcal{N}+1}$

\begin{figure*}[t]
	\centering
	\includegraphics[width=12cm]{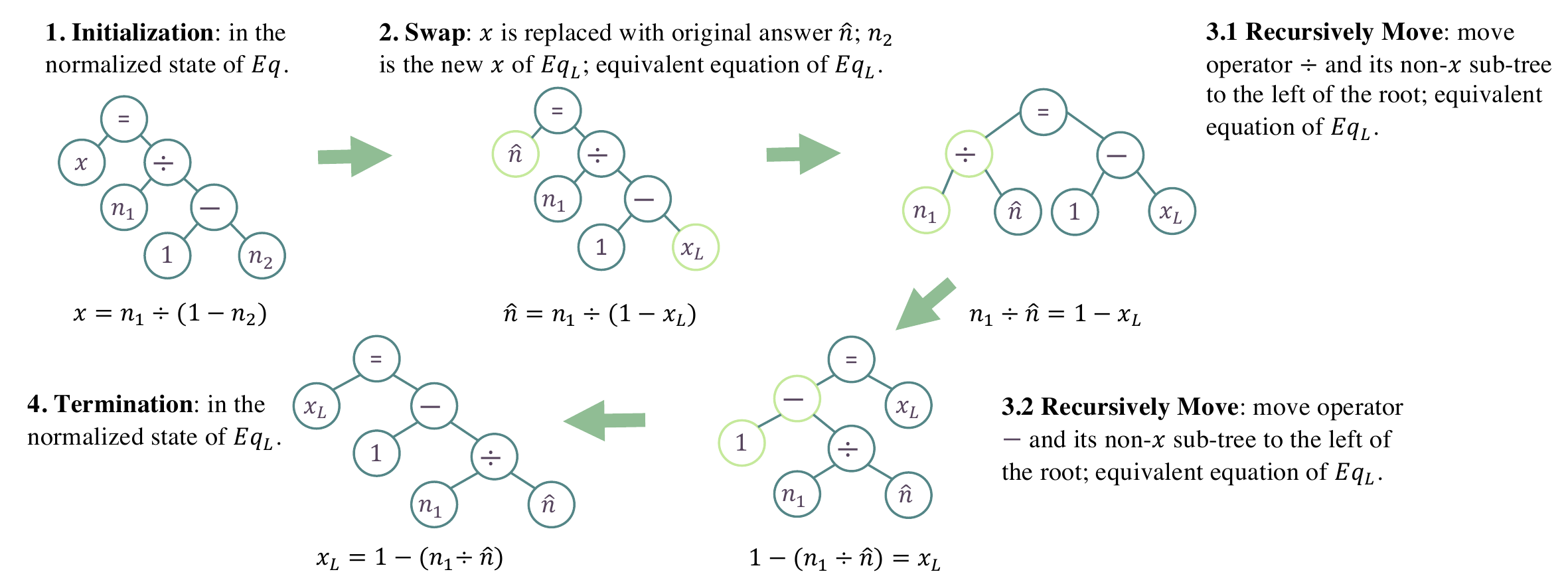}
	\caption{\textbf{The process of generating a new equation $Eq_L$.} 
	\textit{As illustrated in Table \ref{exm_logical}, 
	%$\left | \mathcal{N} \right |=2$ in this case. And 
	the new unknown quantity of $Q_L$ and $Eq_L$ is $n_2$ with $n_1$ and $\hat{n}$ known. %Thus, $Eq_L$ is about the relationship between $n_2$ and $\{n_1, n_3\} \cup \mathcal{C}$. 
	The equation tree in the initialization step is built from $Eq$ in Table \ref{exm_logical}. %From stage 2 to stage 4, the 
	Equation trees from step 2 to step 4 are equivalent equations of $Eq_L$ in different forms. 
	And the tree in the termination step is the normalized form of $Eq_L$.}% and $Eq_L$ could be restored from the tree.}
	}
	\label{fig:equation_generation}
\end{figure*}

\subsubsection{Equation generation with equation tree conversion} 
Since the known and unknown have been changed in the generated question $Q_L$, the corresponding equation $Eq_L$ should be changed accordingly to keep consistency. 
% Besides, it will be hard for neural models to learn if the generated equations are of different forms. 
To normalize the form of equations, the generated equations are transformed to their equivalent equations of a specific form based on mathematical properties with the help of equation trees. 
In the final normalized state, the term $x$ is isolated on the left side of the equation.

Fig. \ref{fig:equation_generation} shows an example of how a consistent and normalized equation $Eq_L$ is generated from the original equation $Eq$. 
\subsubsection{Equation tree} 
Equation tree is built from an equation whose root node is always the equal sign $=$.  
% The leaf nodes are operands including the known and unknown, and the inner nodes are binary operators. 
The left and right sub-trees of the root are expression trees of the left and right sides of an equation, thus the leaf nodes are operands and the inner nodes (except the root) are binary operators. 
% To restore an equation from an equation tree, simply traverse the left and right subtrees of the root node in preorder respectively. % to get two expressions and put the root value $=$ between them.

In the normalized state of an equation tree, the left sub-tree of the root only contains a leaf node $x$. 
% And the rest forms of the equation tree are called un-normalized state. %and need to be transformed to the final state to get equations suitable for neural models.
The same equation tree in different states refers to equivalent equations of different forms. 
To normalize the equation, we transform the equation tree to its normalized state with two main actions based on the addition-subtraction property, the division property, and the multiplication property.
1) Move an operator node $o$ (an inner node) and its non-$x$ sub-tree (the sub-tree of $o$ that does not contain $x$) to another side. 
2) Switch $o$ to its inverse operator, for example, switch $+$ to $-$ or $\times$ to $\div$. 

\subsubsection{The process of generating $Eq_{L}$} 
% The generation process is devided into several stages. 
As shown in Fig \ref{fig:equation_generation}, there are four steps to generate $Eq_L$.
1) Initialization: a normalized equation tree is built from the original $Eq$.
2) Swap: $x$ is substituted for $\hat{n}$, and the quantity $n_i$ which is the new unknown is replaced with $x_L$ (to distinguish with $x$).
Since the new equation tree of $Eq_L$ is not in normalized state, we take the available actions to transform the new equation tree to its normalized state.
3) Recursively Move: All of the operator nodes and their non-$x$ sub-trees are recursively moved to the left of the root node in a top-down manner, leaving only $x_L$ in the right.
% Different actions are taken in different situations, the rules are described as below.
Different actions are chosen in different situations, all based on the natural mathematical properties. 
% \begin{itemize}
% \item When the operator node $o$ is $+$ or $\times$, the first action is taken by making $o$ the left child of the root node. 
% And the new left child of $o$ turns to the original left subtree of the root node while the right child of $o$ is the non-$x$ subtree. 
% The new value of node $o$ is swapped as the second action mentioned.
% When the operator node $o$ is $-$ or $\div$, the location of the $x$ node matters. 
% \item If the $x$ node is in the left subtree of $o$, the process is the same as the above situation. Notice that the value of $o$ should turn from $-$ and $\div$ to $+$ and $\times$ respectively.
% \item If the $x$ node is in the right subtree of $o$, we simply move the node and its non-$x$ subtree to the left without changing its value. 
% The non-$x$ subtree act as the left child of the new node $o$ while the right child of $o$ is the original left subtree of the root node.
% \end{itemize}
% 4) Termination: After moving all nodes except $x_L$ node to the left of the root node, the equation tree is close to the normalized state. 
4) Termination: Simply swapping the left and right sub-trees of the root node will make the equation tree normalized. And $Eq_L$ could be restored from the equation tree.

\section{Experiment}
In this section, we conduct experiments to measure the scale of challenging samples in existing datasets. 
Besides, we evaluate our semantic-based augmentation strategies (denoted as s.based aug. for simplification) with three typical neural MWP solvers to show the improvements of the generalization ability brought in by s.based aug.. 
Extensive ablation studies are also conducted to verify the effectiveness of each augmentation strategy. 

\subsection{Dataset Analysis}
Existing MWP datasets such as AI2 \cite{hosseini2014learning}, SingleEQ \cite{koncel2015parsing}, and AllArith \cite{roy2016unit} only have hundreds of samples. 
While relatively large-scale MWP datasets such as MAWPS \cite{koncel2016mawps} contains very few challenging samples as it is constructed with low lexical and template overlap. %is constructed with low lexical and template overlap, which suggests limited coverage of challenging samples. %amount of challenging samples. 
% As introduced in Sec. \ref{section:intro}, 
% Existing MWP datasets like AI2 \cite{hosseini2014learning} are fairly small, and relatively large-scale MWP dataset MAWPS \cite{koncel2016mawps} contains very few challenging samples as it is constructed with low lexical and template overlap. 
So we conduct our experiments on the largest and most popular MWP dataset Math23k \cite{wang2017deep}, a Chinese MWP dataset with 23,161 pairs of $(Q,Eq)$.

To make the neural models able to deal with discrete tiny local variances, it's necessary for the dataset to contain a great ratio of challenging samples that have similar questions but different equations. 
In this section, we will analyze Math23k from the amount and the quality of challenging samples in the training and the testing set. 

Notably, the analysis in this section are all based on equation templates, which means only the structures of the equations are considered rather than the actual values. 
For example, equations $x=3+2+1$ and $x=5+4+2$ are viewed as the same in template.

\begin{figure}
    \centering
    \begin{tabular}{cc}
       \includegraphics[width=0.5\linewidth]{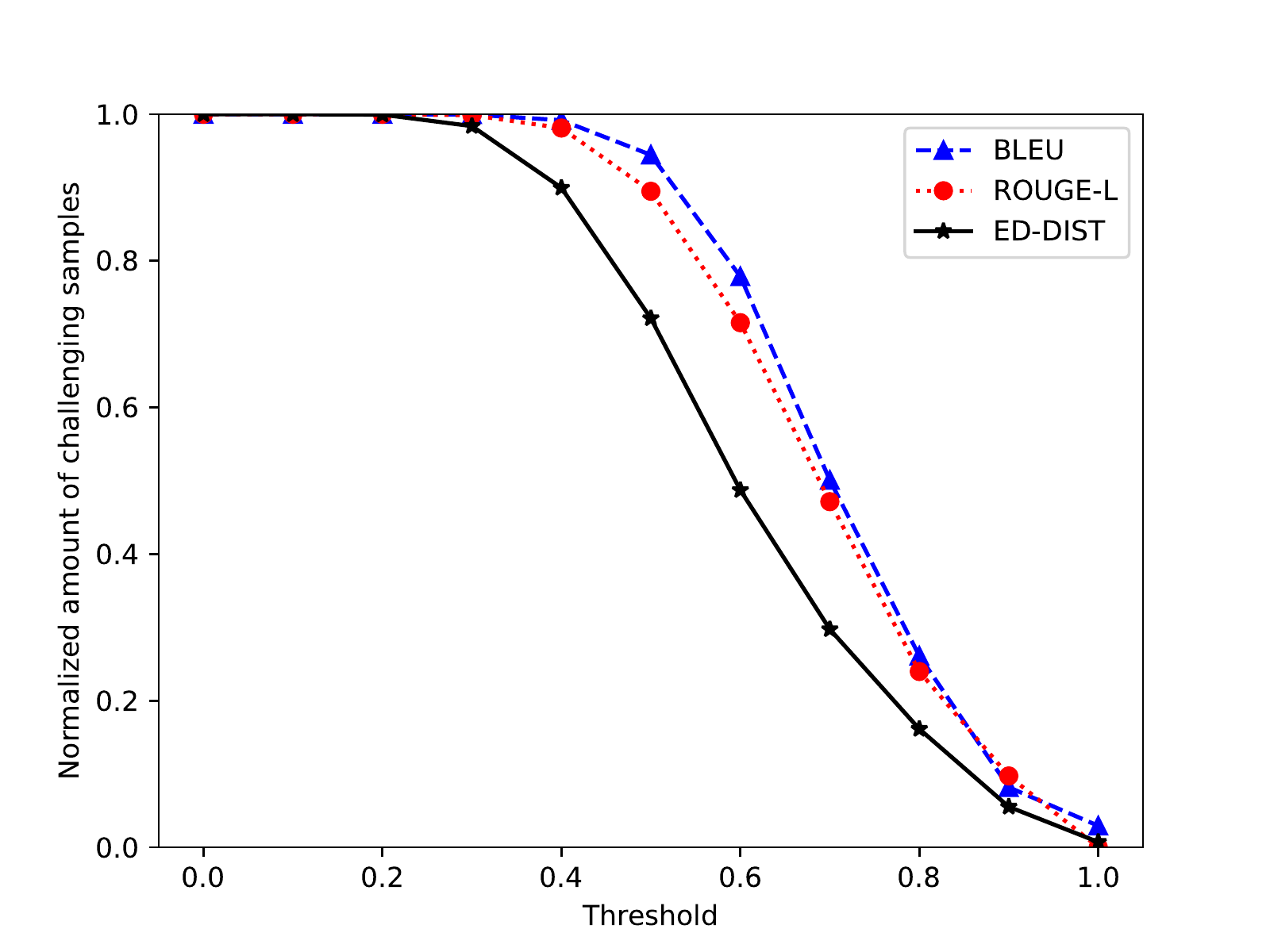}  & \includegraphics[width=0.5\linewidth]{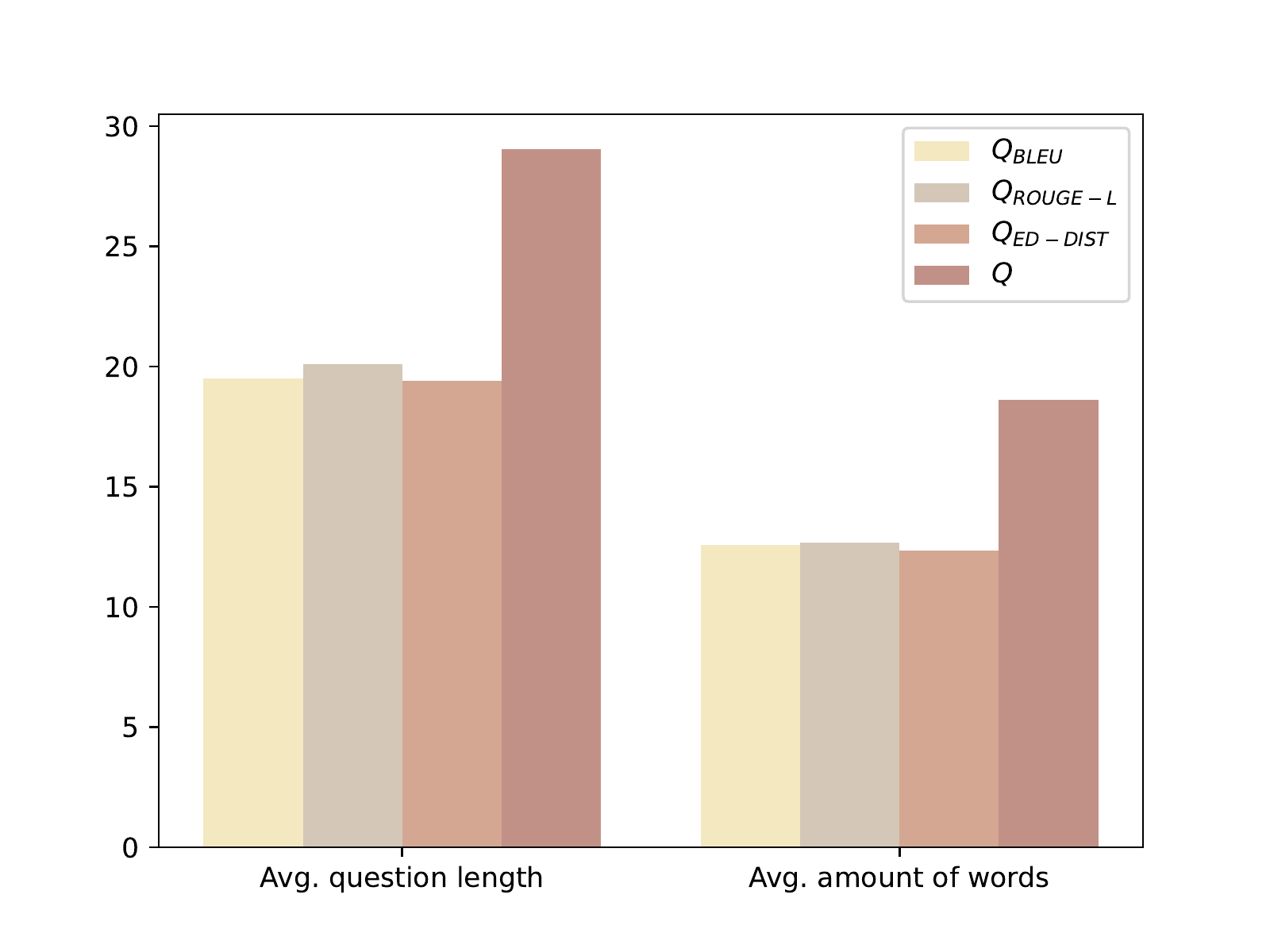} \\
        (a) & (b) 
    \end{tabular}
    \caption{(a) the amount of challenging samples filtered with different similarity score thresholds in the training set. (b) the quality of existing challenging samples.}
    \label{training_set_analysis}
\end{figure}

% \begin{figure}
% 	\centering
% 	\includegraphics[width=0.5\columnwidth]{resource/sample_cnt.pdf}
% 	\caption{The amount of challenging samples filtered with different similarity score thresholds.
% 	% \textit{}
% 	}
% 	\label{training_set_analysis}
% \end{figure}
\subsubsection{The amount of challenging samples in the training set}\label{section:training_set_analysis}
% To explore the amount of challenging samples in the training set $D_{train}$ of Math23k, we calculate textual similarity scores for data pairs that have different equation templates in the $D_{train}$. 
For a question $Q_i \in D_{train}$, if $\exists Q_j \in D_{train}(i \neq j)$ with different equation but similar text, $Q_i$ and $Q_j$ are considered as challenging samples. 
% Data pairs that are different in equations and similar in text are considered as challenging samples. 
% We denote a data pair as $p=(Q_i,Q_j)$, in which $Q_i,Q_j \in D_{train}$ and $Eq_i\neq Eq_j$. 
The similarity of two questions are measured with three similarity scores, namely BLEU, ROUGE-L, and normalized reverse edit distance defined below (referred to as ED-DIST). 
\begin{gather*}
	ED-DIST(Q_i,Q_j)=1-\frac{edit-dist(Q_i,Q_j)}{max(l_{Q_i},l_{Q_j})}
\end{gather*}
$l$ refers to the length of the question and $edit-dist(\cdot)$ is the Levenshtein distance %\cite{levenshtein1966binary} 
of the given pair of questions.
% When $Q_i$ is similar to $Q_j$, $ED-DIST(Q_i, Q_j)$ is closer to 1.
The more similar $Q_i$ and $Q_j$ are, the closer $ED-DIST(Q_i, Q_j)$ is to 1.
Afterwards, we count the amount of challenging samples with the similarity thresholds set to different values as shown in Fig. \ref{training_set_analysis} (a).
The amount of samples in Fig. \ref{training_set_analysis} (a) is normalized with respect to the size of the training set $\left | D_{train} \right |$.
% It's obvious that the amount of challenging samples in Math23k is rather limited as only a few samples meet the condition that $\exists Q^{'}\in D_{train}$ that is similar to it but has different equation (no more than 15\% when the threshold score is set to 0.9).
It's obvious that the amount of challenging samples in Math23k is rather limited as no more than 15\% of the questions meet the condition when the threshold score is 0.9.

% \begin{figure}
% 	\centering
% 	\includegraphics[width=0.5\columnwidth]{resource/sample_quality.pdf}
% 	\caption{The quality of existing challenging samples.}
% 	\label{sample_quality}
% \end{figure}

% \begin{table}
% 	\normalsize
% 	\caption{Examples of existing challenging samples in $D_{train}$.}
% 	\begin{tabular}{m{6.8cm}|m{6.8cm}}
% 		\makecell[c]{$Q_i$} & \makecell[c]{$Q_j$} \\
% 		\hline
% 		The divisor is 8 and the quotient is 2, how about the dividend? & The dividend is 24 and the divisor is 3, how about the quotient? \\
% 		\hline
% 		3 times a number is 300, this number is equal to ? & A number is 7 times 21, this number is equal to ? \\
% 		\hline
% 		Number A is equal to 150, and number B is 20\% more than A. B = ? & Number A is 10.78, and number B is 3 more than B. B = ?
% 	\end{tabular}
% 	\label{cases_of_challenging_samples}
% \end{table}

% \begin{table}
% 	\normalsize
% 	\caption{Examples of manually labeled challenging samples in $D_{test}^{*}$.}
% 	\begin{tabular}{m{1.5cm}<{\centering}|m{6cm}}
% 		% \hline
% 		$Q_i$ & In a parking lot, totally 48 cars and motorcycles are parked. Each car has 4 wheels, and each motorcycle has 3 wheels. If there are 20 motorcycles in the parking lot, how many wheels are there in total? \\
% 		\hline
% 		$Q_j$ & In a parking lot, totally 48 cars and motorcycles are parked. Each car has 4 wheels, and 172 wheels are there in total. If there are 20 motorcycles in the parking lot, how many wheels does a motorcycle have? \\
% 	\end{tabular}
% 	\label{cases_of_additional_test}
% \end{table}

\begin{table}[t]
    \centering
    \caption{Examples of challenging samples.}
    \begin{tabular}{m{2cm}|m{5cm}|m{5cm}}
        \hline
        \multirow{4}{2cm}[-25pt]{challenging samples in $D_{train}$} & \makecell[c]{$Q_i$} & \makecell[c]{$Q_j$} \\
        \cline{2-3}
		& The divisor is 8 and the quotient is 2, how about the dividend? & The dividend is 24 and the divisor is 3, how about the quotient? \\
        \cline{2-3}
		& 3 times a number is 300, this number is equal to ? & A number is 7 times 21, this number is equal to ? \\
		\cline{2-3}
		& Number A is equal to 150, and number B is 20\% more than A. B = ? & Number A is 10.78, and number B is 3 more than B. B = ? \\
		\midrule[1pt]
% 	    {\vspace{-1.2em} challenging samples in $D_{test}^{*}$} & 
        \multirow{1}{2cm}[5pt]{challenging samples in $D_{test}^{*}$} &
    	In a parking lot, totally 48 cars and motorcycles are parked. Each car has 4 wheels, and each motorcycle has 3 wheels. If there are 20 motorcycles in the parking lot, how many wheels are there in total? &
    	In a parking lot, totally 48 cars and motorcycles are parked. Each car has 4 wheels, and 172 wheels are there in total. If there are 20 motorcycles in the parking lot, how many wheels does a motorcycle have? \\
    	\hline
    \end{tabular}
    \label{tab:exm_challenging_samples}
\end{table}

\subsubsection{The quality of the challenging samples}\label{section:quality_of_challenging samples}
% In the above section, we analyze the amount of the possible challenging samples in $D_{train}$ and find that $D_{train}$ contains very limited challenging samples.
High-quality MWP questions are supposed to be longer sentences with rich background descriptions. 
Thus we analyze the average length and words diversity of challenging samples in $D_{train}$.
% We further analyze the quality of the challenging samples in $D_{train}$ by comparing the average length and the words diversity of challenging samples with that of the whole samples in $D_{train}$.
The threshold of similarity scores is set to 0.9 to filter the challenging samples. 
% Questions in $D_{train}$ are considered challenging when $\exists Q^{'} \in D_{train}$ that makes the similarity scores of data pair $p=(Q,Q^{'})$ higher than 0.9.
In Fig. \ref{training_set_analysis} (b), $Q$ refers to all questions in $D_{train}$, while $Q_{metric}$ refers to the challenging samples filtered by certain metrics. e.g. $Q_{BLEU}$ are challenging questions that have another similar question with BLEU score higher than 0.9.
% Questions that have a BLEU score or ROUGE-L score higher than 0.9, or ED-DIST score lower than 0.1 are considered challenging. 
% The average length of these questions and the amount of different words per question are compared with that of the whole training set.
As shown in Fig. \ref{training_set_analysis} (b), the challenging samples filtered by all metrics are much shorter than the average question length , and so is the diversity of words.
Table \ref{tab:exm_challenging_samples} shows some examples of existing challenging samples in $D_{train}$. 
The experimental results indicate that most of the challenging samples in $D_{train}$ are too short and lack background descriptions. 
% the quality of exsiting challenging samples are not quite good.

Since both the amount and quality of existing challenging samples are not quite satisfactory, we propose knowledge and logical augmentation to generate questions with tiny local variances, which brings in massive high-quality challenging samples.

\subsubsection{Testing set analysis}\label{section:testing_set_analysis}
A good testing set requires low lexical and template overlap between the testing and training set \cite{koncel2015parsing}. 
Besides, as the ability to disambiguate tiny local variances in questions largely reflects the generalization ability of neural models, the testing set is supposed to have more challenging samples to better evaluate the ability of neural models.
As shown in Table \ref{analysis_testing_set}, we count the number of equation templates that only appeared in the $D_{test}$ (New eq. template in Table \ref{analysis_testing_set}) to measure the template overlap between the testing and training set. 
Meanwhile, the textual similarity between questions in the testing set and the training set is calculated with ED-DIST as introduced above. Specifically, for each $Q_{te}$ in the testing set, we calculate a similarity score with $s=\max \{ED-DIST(Q_{te},Q_{tr})\}, \forall Q_{tr} \in D_{train}$.
The challenging samples in the testing set are counted as Sec. \ref{section:quality_of_challenging samples} with the threshold set to 0.9.
Results in Table \ref{analysis_testing_set} are normalized with respect to the size of the testing set.

As shown in Table \ref{analysis_testing_set}, the original testing set shares a high template (low new eq. templates) and lexical (high q. similarity) overlap with the training set, and contains limited challenging samples (low num. challenging samples). 
It indicates that the testing set is to some extent similar to the training set, making it hard to evaluate the actual inference ability of neural models perfectly. 
Not to mention the ability of neural models dealing with challenging samples, the limited scope of challenging samples indicates $D_{test}$ is not able to evaluate neural models from this aspect. 

Considering the weaknesses mentioned above, we manually labeled an additional testing set $D_{test}^{*}$ with 380 samples in total which contains a great deal of \textbf{high-quality} challenging samples (Table \ref{tab:exm_challenging_samples} shows a pair of example challenging sample in $D_{test}^{*}$). 
% The quality of both testing sets is measured from several aspects as shown in Table \ref{analysis_testing_set}.
Besides, the additional testing set $D_{test}^{*}$ holds lower lexical and template overlap with $D_{train}$, and also contains more challenging samples as shown in Table \ref{analysis_testing_set}. 
% Thus we believe $D_{test}^{*}$ could better evaluate the generalization ability of neural models than the original $D_{test}$. %and the ability of them dealing with challenging samples.

\begin{table}[t]
	\normalsize
	\caption{Comparison of the two testing sets.}
	\begin{tabular}{m{1cm}<{\centering}m{2cm}<{\centering}m{2cm}<{\centering}|m{2cm}<{\centering}m{2cm}<{\centering}m{2cm}<{\centering}}
	\hline
	\multirow{2}{1cm}{} &\multirow{2}{2cm}{\centering New eq. template} & \multirow{2}{2cm}{\centering Mean q. similarity} & \multicolumn{3}{c}{Num. challenging samples} \\
	& & & BLEU & ROUGE-L & ED-DIST \\
	\hline
	 $D_{test}$ & 0.0401 & 0.694 & 0.0211 & 0.0361 & 0.0201 \\
	 $D_{test}^{*}$ & 0.155 & 0.593 & 0.574 & 0.282 & 0.0421\\
	 \hline
	\end{tabular}
	\label{analysis_testing_set}
\end{table}

\subsection{Experimental Setup}
\subsubsection{Dataset} 
We conduct our experiments on Math23k \cite{wang2017deep}. 
Besides training with the whole training set $D_{train}$, to evaluate the performance of our augmentation approaches on different data sizes, we randomly picked three training sets of sizes 500, 2k, and 20k (the whole training set) from the $D_{train}$ as what have done in \cite{wei2019eda}. 
The neural models are evaluated on both original testing set $D_{test}$ and the manually labeled $D_{test}^*$. 
\subsubsection{Neural model $\mathcal{M}$} 
We evaluate s.based aug. strategies with three most typical MWP neural models. 
\begin{itemize}
\item \textbf{Vanilla seq2seq} (marked as seq2seq). In this paper, we adopt a seq2seq model as \cite{wang2017deep} whose encoder is a 2-layer BiGRU and the decoder is a 2-layer LSTM.
Besides, to help the model learn the slight variation of our augmented data, we utilize the attention mechanism before the feed-forward network.
\begin{gather*}
	\alpha_i = \frac{e^{V\tanh (W_1h_{n_i}+W_2\hat{h_t}+b)}}{\sum_{j=1}^{\left | \mathcal{N} \right|}e^{V\tanh (W_1h_{n_j}+W_2\hat{h_t}+b)}} \\
	h_\mathcal{N} = \sum_{i=1}^{\left | \mathcal{N} \right|}\alpha_{i}h_{n_i} \\
	P(y_t|y_0,...,y_{t-1}) = f(\hat{h_t}\bigoplus h_\mathcal{N})
\end{gather*}
$V, W_1, W_2, b$ are all parameters, and $h_{n_i}$ is the encoder hidden vector of the quantity $n_i$, while $\hat{h_t}$ refers to the decoder hidden vector of the time step $t$.

\item \textbf{GTS} \cite{xie2019goal} is a tree-based neural model which has outperformed previous works significantly. 
\item \textbf{Graph2Tree} \cite{zhang2020graph} is the state-of-the-art neural model for MWP. 
\end{itemize}

\subsubsection{Baseline models for comparison} 
We compare the performances of the three neural models trained with s.based aug. strategies with an extensive set of related work.
\textbf{Math-EN} \cite{wang2018translating} proposed equation normalization based on a vanilla seq2seq model to effectively reduce the target space.
\textbf{TRNN} \cite{wang2019template} applied a seq2seq model to predict a tree-structured template in a bottom-up manner.
\textbf{GROUP-ATT} \cite{li2019modeling} proposed a group attention mechanism to extract intra-relation features.
\textbf{AST-Dec} \cite{liu2019tree} proposed to generate abstract syntax tree of the equation in a top-down manner.

\subsubsection{Baseline augmentation approach} 
We also compare the specifically designed s.based aug. with one of the most popular text augmentation strategy back translation \cite{yu2018qanet}, which is used to generate paraphrase of original sentence. The original sentence is first translated into a pivot language, and back to its original language.
Other augmentation approaches such as synonymy replacement \cite{wei2019eda}, generation-based methods \cite{hu2017toward}, etc. are omitted here due to the intolerable noise for mathematical scenario. 

\subsubsection{Implementation details}
The baseline models are trained with $D_{train}$ while the augmentation models are trained with $D_{train} \cup D_{aug}$. All models are evaluated on the same testing set $D_{test}$. 
To verify the ability of dealing with challenging samples, the three neural models trained with s.based aug. are additionally evaluated on $D_{test}^*$. 
Our evaluation metric is answer accuracy, which is calculated by comparing the answer of the predicted equation and the ground truth one rather than comparing the equation sequence.
The parameters of the vanilla seq2seq model are set as \cite{wang2017deep}: The hidden units of both the encoder and decoder are 512. The word embedding dimension is set to 50 and the dropout for GRU and LSTM are set to 0.5. 
The number of epochs and mini-batch size are 80 and 32 respectively. We adopt an early stop policy after the accuracy of the validation set not increasing for 10 epochs.
Adam optimizer \cite{kingma2014adam} is used with learning rate set to 0.001, $\beta_1=0.9$ and $\beta_2=0.999$, and the learning rate is halved every 10 epochs.
Settings of GTS and Graph2Tree are the same as that in \cite{xie2019goal} and \cite{zhang2020graph}.

\begin{table}[t]
    \centering
    \caption{\textbf{Answer accuracy} (\%) of neural models evaluated on the two testing sets. We evaluate s.based aug. on three most typical neural models. The improved performances w/ augmentation are shown in bold.}
    \begin{tabular}{l|c|c|c|c|c|c|c}
        \toprule
        & \multicolumn{3}{c|}{$D_{test}$} &  \multicolumn{4}{c}{$D_{test}^{*}$} \\
        \hline
        &  & & & \multicolumn{2}{c|}{All samples} & \multicolumn{2}{c}{Challenging samples} \\
        \hline
        \multirow{2}{*}{\textbf{Model}} & w/o & w/ & w/ & w/o & w/ & w/o & w/ \\
			 & aug. & s.based aug. & back trans. & aug. & s.based aug. & aug. & s.based aug.\\
        \hline
        Math-EN & 66.7 & - & - & - & - & - & -\\
        TRNN & 66.9 & - & -  & - & - & - & - \\
        AST-Dec & 69.0 & - & -  & - & - & - & - \\
        GROUP-ATT & 69.5 & - & -  & - & - & - & - \\
        \hline
        seq2seq & 66.1 & \textbf{71.2} \scriptsize{$\uparrow$5.1} & \textbf{66.4} \scriptsize{$\uparrow$0.3} & 30.0 & \textbf{53.2} \scriptsize{$\uparrow$\textcolor{red}{23.2}} & 11.4 & \textbf{27.3} \scriptsize{$\uparrow$ \textcolor{red}{15.9}} \\
        \hline
        GTS & 75.6 & \textbf{76.1} \scriptsize{$\uparrow$0.5} & 75.5 \scriptsize{$\downarrow$0.1} & 33.2 & \textbf{53.2} \scriptsize{$\uparrow$\textcolor{red}{20.0}} & 11.4 & \textbf{36.4} \scriptsize{$\uparrow$ \textcolor{red}{25.0}} \\
        \hline
        Graph2Tree & 77.4 & 77.0 \scriptsize{$\downarrow$0.4} & 76.9 \scriptsize{$\downarrow$0.5} & 30.3 & \textbf{52.4} \scriptsize{$\uparrow$\textcolor{red}{22.1}}& 11.4 & \textbf{40.9} \scriptsize{$\uparrow$ \textcolor{red}{29.5}} \\
        \bottomrule
    \end{tabular}
    \label{model_comparision}
\end{table}

\subsection{Results}
\subsubsection{Comparison results}\label{section:comparison_results}
Table \ref{model_comparision} shows the answer accuracy of neural models trained with s.based aug. strategies compared with back translation and various baseline models. 
It's obvious that specifically designed s.based.aug. performs much better than back translation on MWP task. Since MWP has a strict requirement of precision, the noises brought in by back translation are sometimes unacceptable. We notice that many of the generated questions by the means of back translation are inconsistent with their equations.
Trained with s.based aug., the accuracy of the seq2seq model increases by 5.1\%, which is able to beat many other complex neural models with 71.2\% answer accuracy.
% The benefits of augmentation on the other two models GTS and Graph2Tree are not obvious. 
Performance gains of simple neural networks like seq2seq are much more than that of complex neural models like GTS and Graph2Tree. 
Our explanation is that complex models have better inference ability, making them able to learn more useful features from limited data, while simpler models have to learn these features from more diverse data.
However, as analysed in Sec. \ref{section:testing_set_analysis}, considering the original testing set $D_{test}$ holds a high overlap with the training set, 
we reasonably suspect that the performance decrease is caused by the complex models likely to be overfitting and learns some dataset-specific features before augmentation. 
And the massive challenging samples brought in by our augmentation strategies may confuse the neural models, since none of them have specially designed to handle the tiny local variances.
We'll further verify our hypothesis in Sec. \ref{section:case_study}.
% We further conduct an additional evaluation on a manually labeled testing set $D_{test}^{*}$ which holds a lower overlap with the training set (\ref{section:testing_set_analysis}). 
% \subsubsection{Additional evaluation results}

As described in Sec. \ref{section:testing_set_analysis}, the manually labeled testing set $D_{test}^*$ holds lower overlap with the training set and contains 380 samples with a significant ratio of challenging samples.
The three neural models trained w/ and w/o s.based aug. are tested on $D_{test}^{*}$ to further evaluate the generalization ability of them.
As shown in Table \ref{model_comparision}, we calculate the answer accuracy for all the questions and the challenging samples respectively. 
The challenging samples are filtered as \ref{section:quality_of_challenging samples} with the threshold set to 0.9. 
For the accuracy of the challenging samples, a challenging sample $Q_i$ is viewed to be correctly answered only if all its similar questions $Q_j$ has been correctly answered. 
% For the accuracy of the challenging samples, as what have done in Sec. \ref{section:quality_of_challenging samples}, data pairs that have a similarity score higher than 0.9 are viewed as challenging samples.
% Notably, if data pairs $p=(Q_i,Q_j)$ and $p=(Q_j,Q_m)$ are both challenging samples, then we combine them to a data set which is $S=\{Q_i,Q_j,Q_m\}$.
% Afterwards we have 44 challenging data sets in total. 
% And only if the neural model correctly predict the answers of all the questions in a challenging data set, we count it as one hit.

% Since the additional testing set $D_{test}^{*}$ holds a lower lexical and template overlap with $D_{train}$ than the original testing set $D_{test}$, the preformance on the additional testing set $D_{test}^{*}$ is more appropriate to show the actual generalization ability of the neural models. And the significant performance improvement indicates that our augmentation approaches are able to improve the generalization ability of the neural models.
% Besides, the results of the answer accuracy for the challenging samples also indicates that our augmentation methods largely help the neural models to learn such tiny local variances in MWP.

As shown in Table \ref{model_comparision}, s.based aug. strategies significantly boost the performances of all the three neural models with more than 20\% increments. Besides, the ability of neural models dealing with discrete tiny local variances has also been largely improved. The results suggest that the s.based aug. strategies successfully benefit the generalization ability of the neural models. 
However, since existing MWP neural models hardly considered the discrete local variances which lead to respectively low accuracy, there is still a large space for future work to improve the ability of neural models dealing with such challenges.

% Besides, we will analyze the error cases in Sec. \ref{section:case_study} to verify our assumptions about the performance decrease.

\begin{table}[t]
	\normalsize
    \centering
	\caption{The ablation study on datasets with different sizes. 
	The best performance on each dataset is shown in bold.}
        \begin{tabular}{l|c c c}
            \hline
             & \multicolumn{3}{c}{\textbf{Training Set Size}} \\
            \textbf{Model} & 500 & 2k & 20k \\
            \hline
            seq2seq & 9.52 & 31.6 & 66.1 \\
            \hline
            seq2seq \\
            +knowledge & 12.3 \scriptsize{$\uparrow$2.78} & 33.5 \scriptsize{$\uparrow$1.90} & 67.0 \scriptsize{$\uparrow$0.90}\\
            \hline
            seq2seq \\
            +logic & 13.0 \scriptsize{$\uparrow$3.48} & 34.1 \scriptsize{$\uparrow$2.50} & 67.3 \scriptsize{$\uparrow$1.20}\\
            \hline
            seq2seq \\
            +s.based aug. & \textbf{16.7} \scriptsize{$\uparrow$7.18} & \textbf{36.0} \scriptsize{$\uparrow$4.40} & \textbf{71.2} \scriptsize{$\uparrow$5.10}\\
            \hline
        \end{tabular}
        \label{data_size_ablation}
\end{table}

% the amount of data generation
% \begin{figure}[t]
%     \small
% 	\centering
% 	\includegraphics[width=0.9\columnwidth]{resource/generate_amount_experiment.pdf}
% 	\caption{The performance gains of s.based aug. with varied generated ratios.}
% 	\label{generate_ratio}
% \end{figure}

\subsubsection{Ablation study} 

As illustrated in Tabel \ref{data_size_ablation}, both knowledge and logic guided augmentation methods contribute to the performance gains. 
And logical augmentation performs better than knowledge augmentation on all three datasets. This gap is more obvious on smaller datasets.
We guess that on smaller datasets, the lack of problem types is more severe, thus enriching the equation templates is more needed.
Besides, the results have shown that smaller datasets benefit more from the augmentation strategies than larger ones.

\begin{table*}[t]
% 	\normalsize
    \small
	\caption{Examples of error cases that are predicted wrongly after augmentation.}
	% \begin{tabular}{m{2.1cm}<{\centering}|m{2.4cm}<{\centering}}
	\begin{tabular}{m{3.5cm}|m{3.5cm}|m{2.5cm}|m{2.5cm}}
	    \hline
		 \makebox[3.5cm][c]{$Q_{tr}\in D_{train}$} & \makebox[3.5cm][c]{$Q_{te} \in D_{test}$} & \makebox[2.5cm][c]{pred. $Eq$ of $Q_{te}$} & \makebox[2.5cm][c]{tgt. $Eq$ of $Q_{te}$}\\
		\hline
		 A dictionary is priced at $n_1$ yuan, and after $n_2$ of the sale, the price is still $n_3$ higher than the purchase price. The purchase price of this dictionary is ?
		 & A dictionary is priced at $n_1$ yuan, and after $n_2$ of the sale, it will earn $n_3$. The purchase price of this dictionary is ?
		& $x=n_1\times(1+n_3)\div n_2$
		& $x=n_1\times n_2\div (1+n_3)$\\
		% 除数是8，商是2，被除数=．& 被除数是24，除数是3，商=？\\
		\hline
		A project can be completed in $n_1$ days if $n_2$ people come to do it. If $n_3$ people do it, how many days can it be done? 
		& $n_1$ workers will complete a project within $n_2$ days. If it takes $n_3$ days to complete, how many workers are needed?
		& $x=1\div (n_1\times n_2) \div n_3$
		& $x=(n_1\times n_2) \div n_3$\\
		% 一个数的3倍是300，这个数=．& 一个数是21的7倍，这个数=．\\ 
% 		\hline
% 		The original staff of a certain agency is $n_1$, and there are currently $n_2$ staff. How many percent of the staff are reduced compared to the original staff?
% 		& The original staff of a certain agency is $n_1$, and there are currently $n_2$. How many percent does it reduce?
% 		& \centering $x=n_2\div n_1$
% 		& $x=n_1-n_2 \div n_1$\\
        \hline
		% 甲数是150，乙数比甲数多20\%，乙数=． & 甲数是10.78，乙数比甲数多3.88，乙数=？
	\end{tabular}
	\label{case_study_seq2tree}
\end{table*}

\subsection{Case study}\label{section:case_study}
% analyza samples: baseline correct, augmentation wrong
In this section, we'll analyze the questions that are predicted correctly before the neural models trained with augmented data but are predicted wrongly afterwards. 
For each error case, we search for a most similar question in $D_{train}$ to see whether these problems have occurred during training phase. 
And it turns out that most of the error cases have a nearly the same question in $D_{train}$ as shown in Table \ref{case_study_seq2tree}.
The $Q_{te}\in D_{test}$ are error predicted questions in the testing set. $Q_{tr} \in D_{train}$ are their similar questions found in the training set. %And pred. $Eq$ is the mistaken prediction result after augmentation. 
% As shown in the table, both the textual description and the problem types of the questions in the training set are similar to the error cases.
In the first row of Table \ref{case_study_seq2tree}, the target $Eq$ of the $Q_{tr}$ is $x=n_1\times n_2\div (1+n_3)$. 
% And one of the equation of the augmented samples is $x=(n_4 \times (1+n_3))\div n_2$ as described in \ref{section:logical_augmentation}, which is equivalent to the mistaken predicted equation in template.
According to the logical augmentation described in Sec. \ref{section:logical_augmentation}, one of the $Eq_L$ could be $(\hat{n}\times(1+n_3))\div n_2$, which is equivalent to the mistakenly predicted equation in template.
This result further support our hypothesis in Sec. \ref{section:comparison_results} and explain the performance decreases of complex models on $D_{test}$.
% Considering none of the existing neural MWP solvers have been designed to deal with the local variances, we doubt that they have limited ability to handle the challenging samples in MWP.
% Thus, the introduced challenging samples may hurt the overfitting on the training set leading to the decreases of performances on $D_{test}$. 
% The results on $D_{test}^*$ show that the performances on fresh samples are largely improved with s.based aug., reflecting the improvement of the generalization ability of neural models trained with augmented data.
% It indicates that correctly answering these questions may not reflect the inference and generalization ability of the neural models. 
% The neural models may be overfitting on the training set before augmentation, and the massive challenging samples with only tiny local variances but different labels brought in by the augmentation strategies may break the overfitting on the original training set.

% \section{Case Study}\label{section:case_study}
% \input{6.case_study.tex}

\section{Conclusion \& Discussion}
In this paper, we argue that discrete tiny local variances are a big challenge for neural models which previous works have ignored. 
And we propose a set of novel semantic-based data augmentation methods to supplement existing datasets with challenging samples. %involve knowledge guided entity replacement and logic guided problem reorganization. 
Both augmentation methods we proposed are able to generate coherent questions with consistent labels and largely diversify both textual descriptions and equation templates. 
% Extensive experiments are conducted, and the results have shown that exsiting datasets lack the kind of samples provided by our methods. 
% Besides, the augmentation approaches are able to boost the generalization ability of neural models and also the ability of dealing with challenging samples.
Extensive experimental results have shown the necessity and effectiveness of the approaches we proposed. 
Besides, the idea we proposed could also be transferred to other math-related tasks like MWP generation.
% improving the generalization ability of neural MWP solvers with our data augmentation methods,  

% \section{Acknowledgement}
% This work
% was supported by National Key Research and Development Project
% (No.2020AAA0109302), Shanghai Science and Technology Innovation
% Action Plan (No.19511120400) and Shanghai Municipal Science
% and Technology Major Project (No.2021SHZDZX0103).

% ---- Bibliography ----
%
% BibTeX users should specify bibliography style 'splncs04'.
% References will then be sorted and formatted in the correct style.
%
\bibliographystyle{splncs04}
\bibliography{citation}

\end{document}